\documentclass[runningheads]{llncs}

\usepackage{eccv}

\usepackage{eccvabbrv}

\usepackage{graphicx}
\usepackage{booktabs}
\usepackage{marvosym}
\usepackage{multirow}
\usepackage{enumitem}
\usepackage{makecell}
\usepackage{wrapfig}
\usepackage{caption}
\usepackage{colortbl}
\usepackage[accsupp]{axessibility}  

\usepackage{hyperref}

\usepackage{orcidlink}

\begin{document}

\title{Prediction Exposes Your Face: Black-box Model Inversion via Prediction Alignment} 

\titlerunning{Black-box Model Inversion via Prediction Alignment}

\author{Yufan Liu\inst{1,2} \and
Wanqian Zhang\inst{1}\textsuperscript{\Letter} \and
Dayan Wu\inst{1} \and
Zheng Lin\inst{1,2} \and
Jingzi Gu\inst{1} \and
Weiping Wang\inst{1,2}}

\authorrunning{Y. Liu et al.}

\institute{Institute of Information Engineering, Chinese Academy of Sciences \and
School of Cyber Security, University of Chinese Academy of Sciences
\email{\{liuyufan,zhangwanqian,wudayan,linzheng,gujingzi,wangweiping\}@iie.ac.cn}\\
}

\maketitle\let\thefootnote\relax\footnotetext{\textsuperscript{\Letter}Corresponding author}

\begin{abstract}
Model inversion (MI) attack reconstructs the private training data of a target model given its output, posing a significant threat to deep learning models and data privacy. 
On one hand, most of existing MI methods focus on searching for latent codes to represent the target identity, yet this iterative optimization-based scheme consumes a huge number of queries to the target model, making it unrealistic especially in black-box scenario. 
On the other hand, some training-based methods launch an attack through a single forward inference, whereas failing to directly learn high-level mappings from prediction vectors to images.
Addressing these limitations, we propose a novel Prediction-to-Image (P2I) method for black-box MI attack. 
Specifically, we introduce the Prediction Alignment Encoder to map the target model's output prediction into the latent code of StyleGAN. 
In this way, prediction vector space can be well aligned with the more disentangled latent space, thus establishing a connection between prediction vectors and the semantic facial features. 
During the attack phase, we further design the Aligned Ensemble Attack scheme to integrate complementary facial attributes of target identity for better reconstruction.
Experimental results show that our method outperforms other SOTAs, e.g., compared with RLB-MI, our method improves attack accuracy by 8.5\% and reduces query numbers by 99\% on dataset CelebA.
\end{abstract}

\section{Introduction}
\label{sec:intro}
Deep neural networks (DNNs) have been widely applied in various scenarios such as finance, healthcare and autonomous driving. 
Despite the great success on downstream tasks, the collection of DNNs' training data inevitably involve private and sensitive personal information. 
Malicious people can launch various attacks on DNNs to steal users' private information \cite{rigaki2020survey}, which may pose serious threats to user privacy\cite{fredrikson2015model,shokri2017membership,tramer2016stealing}, especially on sensitive information like face images. 
In this paper, we focus on the model inversion (MI) attack, a representative privacy attack paradigm that reconstructs the training data of target model.
Specifically, once obtaining the access to target model and the output predictions, the adversary can attack a face recognition system to reconstruct sensitive face images.
\begin{figure}[t]
    \centering
    \includegraphics[width=0.6\columnwidth]{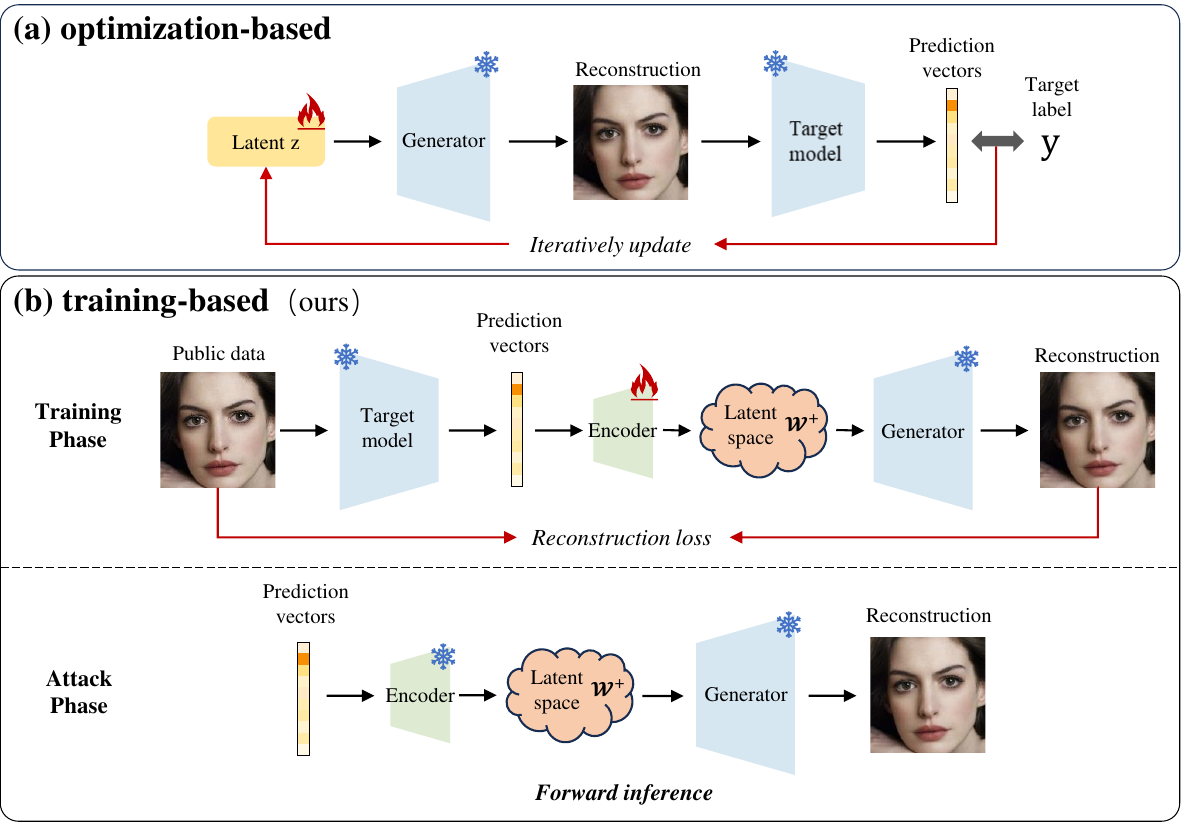}
    \caption{
    Previous optimization-based methods iteratively update latent vector z within a fixed prior generator, involving enormous query numbers to target model.
    Differently, our method works in a training-based manner, optimizing a prediction-to-image inversion model and reconstructing face images through a simple forward inference.
    }
    \label{fig:1}
\end{figure}

Having access to target model's parameters, white-box MI attacks \cite{zhang2020secret,chen2021knowledge,yuan2023pseudo} recover high-fidelity private images by searching the latent space of generative networks. 
While in black-box scenario, where only the output prediction scores are available, the adversaries typically adopt genetic algorithm \cite{an2022mirror} or reinforcement learning \cite{han2023reinforcement} to find the optimal latent vector.
In the most challenging label-only scenario \cite{kahla2022label}, the attackers only have predicted hard labels, thus resort to estimation of the gradient direction away from decision boundaries. 
Despite these diversities, common and central to them is an optimization procedure, i.e., searching the input space to find the exact feature value with maximum likelihood under the target model.

However, as in Fig. \ref{fig:1}, this optimization-based paradigm updates the input vector in an iterative way, which inevitably involves enormous queries to target model.
This is impractical in real-world especially the query-limited scenarios like online machine learning services such as Amazon Rekognition and Google’s cloud vision API. 
For instance, RLB-MI~\cite{han2023reinforcement} takes 40,000 epochs (nearly 120k queries) to reconstruct only one identity.
On the other hand, Yang \etal \cite{DBLP:conf/ccs/YangZCL19} consider the target model as an encoder and train a decoder separately to reconstruct images. 
This vanilla training scheme introduces a simple-structured model, which fails to directly learn high-level mappings from prediction vectors to images. 
In other words, it is difficult for a shallow inversion model, usually 3-4 convolutional layers, to provide disentangled image features, thus leading to unsatisfactory attack performance.

In this paper, we propose the Prediction-to-Image (P2I) inversion model, generating target identity's images through a simple forward inference.
As in Fig. \ref{fig:1}, previous optimization-based methods perform the iterative search for high-likelihood reconstruction from the target model, which could be particularly challenging for high-dimensional search space.
Differently, our method reconstructs face images directly in a generative and disentangled manner.
Specifically, the P2I model consists of a prediction alignment encoder and a StyleGAN generator, which maps the predictions to the StyleGAN's $\mathcal{W}^+$ space and to the image space.
This \textit{prediction-$\mathcal{W}^+$-image} scheme successfully aligns the prediction vector space with the disentangled $\mathcal{W}^+$ space. 
Subsequently, this alignment provides connections between the prediction vector space and image space, contributing to semantically continuous face embeddings on the target identity. 
Besides, to provide more prior knowledge for training, we formulate the training set by public images with highest probabilities for each target identity. 
This public data selection implicitly preserves the facial attribute overlaps between public and private images, preserving rich semantic information for the P2I model.
During the attack phase, we further propose the aligned ensemble attack, integrating public images' latent codes of $\mathcal{W}^+$ space and utilizing the contained target’s facial attributes for better reconstruction.
Empirically, our method shows a significant boost in black-box MI attack accuracy, visual quality and reduced query numbers.

Our contributions can be summarized as follows:

\setlist[itemize]{noitemsep,leftmargin=*,topsep=0em}
\begin{itemize}[leftmargin=6mm]
    \item 
    We propose a novel Prediction-to-Image (P2I) method for black-box model inversion attack. 
    By integrating the proposed prediction alignment encoder with StyleGAN's generator, P2I aligns the prediction vector space with the disentangled $\mathcal{W}^+$ space, providing semantically continuous face embeddings for the target identity.
    \item 
    We propose the aligned ensemble attack scheme to incorporate rich and complementary facial attributes of public images within the $\mathcal{W}^+$ space, further improving the inversion performance.
    \item 
    Extensive experiments show the effectiveness of our method compared with other baselines across various settings like different target models and target datasets, indicating the superiority of our method.
\end{itemize}

\section{Related Works}
\label{sec:formatting}
\subsection{Model Inversion Attack} 
Model Inversion (MI) attacks leverage the target model's output to reconstruct the training data, putting machine learning models at risk of data privacy leakage. 
Fredrikson \etal \cite{DBLP:conf/uss/FredriksonLJLPR14} first propose the MI attack on pharmacogenetic privacy issues, which however easily sticks into local minima due to direct optimization in the high-dimensional image space~\cite{fredrikson2015model}.  
Recent works \cite{zhang2020secret,chen2021knowledge,wang2021variational,DBLP:conf/icml/StruppekHCAK22,nguyen2024label,yuan2022secretgen,kahla2022label,an2022mirror,yuan2023pseudo,han2023reinforcement,nguyen2023re,ye2023c2fmi} incorporate GANs into the pipeline and optimize GAN's latent space instead of image space.
Yuan \etal \cite{yuan2023pseudo} introduce conditional GAN to find a fixed search space for each category, greatly narrowing down the search space. 
Han \etal \cite{han2023reinforcement} focus on black-box MI attack and formalize the latent space search as a Markov Decision Process (MDP) problem.
Kahla \etal \cite{kahla2022label} first explore the label-only MI attack and sample multiple points to estimate the gradient of a random vector.
Nguyen \etal ~\cite{nguyen2023re} study the issues of optimization objective and overfitting for a generic performance boost of all MI algorithms.
However, these methods need enormous iterations and queries to find a latent code when each single attack, which is obviously time-consuming and unrealistic.

In contrast to optimization-based methods, Yang \etal \cite{DBLP:conf/ccs/YangZCL19} first propose a training-based approach. 
These methods \cite{DBLP:conf/ccs/YangZCL19,zhu2022label,ye2022model} generally requires training an additional inversion model, using the output of the target model as input and the images as output. 
During the attack phase, attackers only need to input the target model's output representing the target identity, and then reconstruct the image through one forward propagation. 

\subsection{GAN Inversion} 
The goal of GAN inversion is to encode a given image into the GAN's latent space, and then invert the latent code to obtain the reconstructed high-fidelity image. 
GAN inversion is broadly categorized into three types \cite{xia2022gan}: optimization-based, encoder-based, and the hybrid. 
The optimization-based methods \cite{creswell2018inverting,abdal2019image2stylegan,abdal2020image2stylegan++,bau2020semantic,gu2020image} minimize the error between input image and reconstructed image with gradient descent, while the encoder-based ones \cite{collins2020editing,kang2021gan,alaluf2021restyle,tov2021designing,richardson2021encoding,yao2022style,wang2022high,liu2023delving} train an encoder to map the real image into the latent space as latent code, and the inversion can be achieved by a one-time forward propagation. 
The hybrid method \cite{zhu2020domain} first trains the encoder to get the latent code and then optimizes it later.
Notably, the premise of GAN inversion is that the performer has a real image, which is essential for the task of image reconstruction.
In MI attack, however, we aim to utilize the target model's output prediction to reconstruct a representative image of the target identity in the training dataset.

\section{Method}
\subsection{Problem Formulation}
\textbf{Attacker's goal.}
Consider a target model $T:\mathcal{X} \rightarrow{\mathcal{P}}$ mapping from image space $\mathcal{X}$ to prediction vector space $\mathcal{P}$, which is trained on a private dataset $D_{priv} = {\{(\tilde{x}_i, \tilde{p}_i)\}}_{i=1}^{N_{priv}}$, where $N_{priv}$ is the total number of private samples, $\tilde{x}_i\in{\mathbb{R}^{d}}$ is the input image and $\tilde{p}_i\in{\mathbb{R}}^C$ is the corresponding prediction vector, $C$ is the total number of classes. 
In this work, similar as in \cite{han2023reinforcement}, the target model is specified as the face recognition model, and the attacker aims to recover a representative face image of a given identity.
Formally, the objective of our method is to learn an inversion model that can correctly map the output prediction to its corresponding target identity's image. 

\noindent\textbf{Attacker’s knowledge.}
Our work focuses on the black-box scenario where the attacker does not know neither the internal structure nor the model parameters, yet can only obtain the model's output predictions, i.e., the confidence scores for each class. 
Though the attacker has no access to the private dataset, it is reasonable to assume that he knows what task the model performs, and can easily collect task-related public dataset $D_{pub}$ from the Internet for training~\cite{zhang2020secret,yuan2023pseudo,han2023reinforcement}.
Note that there is no identity overlaps between the public and private datasets.

\begin{figure}[t]
  \centering
    \includegraphics[width=\linewidth]{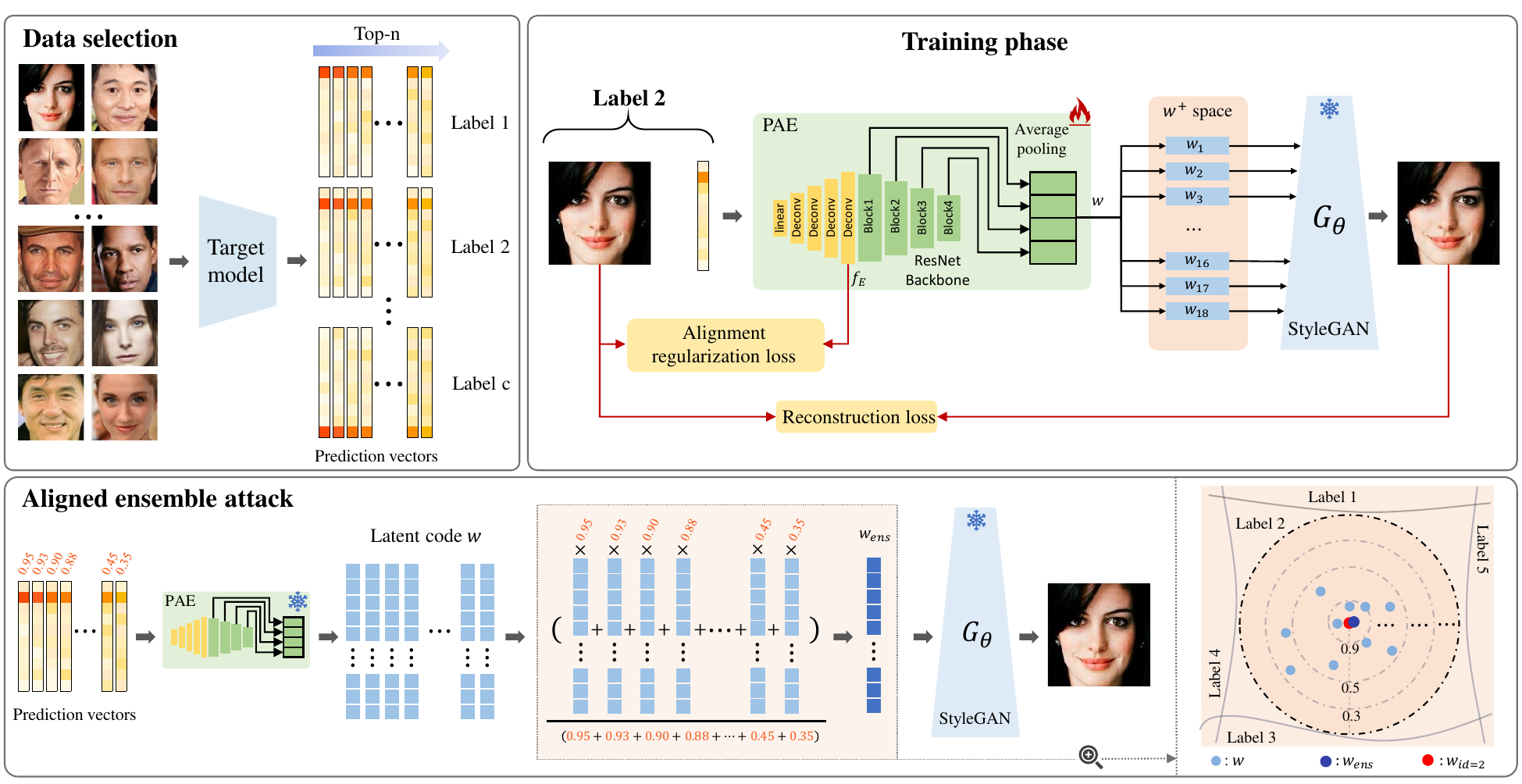}
    \caption{
    Overall pipeline of P2I method. 
    We first form training data by selecting top-$n$ public images with highest confidence for each identity. 
    The Prediction Alignment Encoder (PAE) maps prediction vectors into the latent code of disentangled $\mathcal{W}^+$ space, which are then fed into the fixed StyleGAN's generator to reconstruct high-fidelity target image.
    Furthermore, we introduce aligned ensemble attack to integrate different $w$, which essentially aims to find the centroid $w_{ens}$ and make it closer to the target identity's $w_{id}$, contributing to better attack performance. 
    }
    \label{fig:2}
\end{figure}

\begin{figure*}[t]
  \centering
  \includegraphics[width=\linewidth]{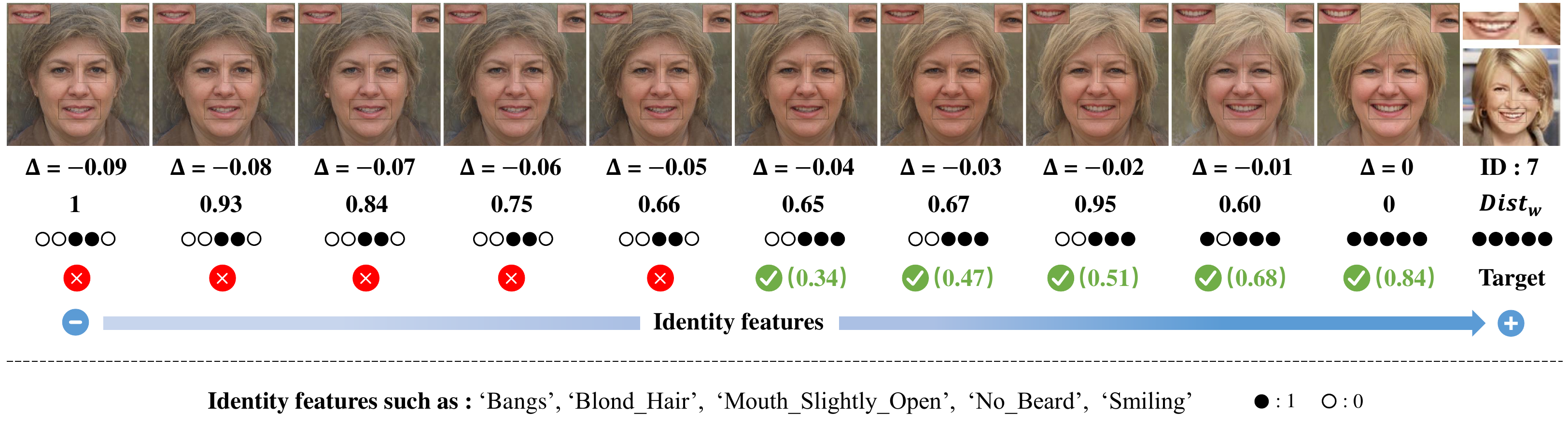}
   \caption{
   Visualizations of the interpolation on prediction vectors along the target dimension.
   As prediction value increases, reconstructed image gradually approaches target visual appearance.
   This is consistent with the decreasing normalized distance $Dist_w$ between the latent codes of target and reconstructed image.
   Besides, results on facial attribute classifications and identity recognition (especially the zoom-in parts of mouth and eye) also justify the \textit{prediction-$\mathcal{W}^+$-image} alignment. 
   }
   \label{fig:3}
\end{figure*}

\subsection{Prediction Alignment Encoder}
Fig. \ref{fig:2} shows the overall framework of our method.
We first form the training set by publicly collected top-$n$ images with highest probabilities for each identity. 
During the training phase, Prediction Alignment Encoder maps prediction vector into disentangled $\mathcal{W}^+$ space, followed by the fixed StyleGAN's generator to reconstruct high-fidelity target image.
This \textit{prediction-$\mathcal{W}^+$-image} scheme establishes connections between prediction vector space and the image space, providing semantically continuous face embeddings on the target identity.

Recently, as one of the excellent works on GAN inversions, \cite{yao2022style} utilizes StyleGAN to reconstruct images by representing visual attributes with different latent dimensions within the disentangled $\mathcal{W}^+$ space. 
Here, motivated by this, we raise the following questions for MI attack: 
Instead of the optimization-based paradigm with high cost and low efficiency,
\textit{Can we} directly train an inversion model to reconstruct images of any given identity through a simple forward inference?
\textit{Can we} further align the prediction vector space with the $\mathcal{W}^+$ space, connecting between prediction vectors and disentangled facial attributes?

Thus, we propose the Prediction-to-Image inversion model, consisting of a Prediction Alignment Encoder $E$ and a StyleGAN generator $G$. 
Specifically, given the input prediction $p$ of image $x$, the Prediction Alignment Encoder aims to learn the mapping $E:\mathcal{P}\rightarrow{\mathcal{W}^+}$, where $p\in{\mathcal{P}}$ and $w\in{\mathcal{W}^+}$, such that the output $G{(E(p))\approx{x}}$.
Concretely, prediction vector will first pass through one linear layer and several deconvolution layers, generating a feature $f_E$ with the same size of image. 
Next, it will go through a ResNet backbone and the output features from each block are connected and flattened into a tensor of $1\times8,640$ size after the average pooling layer. 
Then, it is mapped to a $1\times18\times512$ latent code in $\mathcal{W}^+\subset{\mathbb{R}^{18\times512}}$ space through $18$ parallel linear layers, which is a more disentangled latent space contributing to style mixing \cite{karras2019style,richardson2021encoding} and image inversion \cite{creswell2018inverting,abdal2019image2stylegan,abdal2020image2stylegan++,richardson2021encoding}.
Subsequently, the fixed StyleGAN generator $G$ preserves the capability of generating high-resolution images with various styles and stochastic details. 
By integrating the proposed $E$ and $G$, our method aligns the prediction vector space with the disentangled $\mathcal{W}^+$ space, providing semantically continuous face embeddings for the target identity.

To justify this \textit{prediction-$\mathcal{W}^+$-image} alignment achieved by our method, we report empirical visualizations in Fig. \ref{fig:3}. 
Concretely, we select one target private image, interpolate prediction vectors of public images (classified as the target) along the target dimension (maintaining sum of the vector as $1$), and show corresponding reconstructed images.
Clearly, as the value on target dimension increases, the reconstructed images gradually approach the target image's visual appearance, indicating the alignment between predictions and reconstructed images.
This is also consistent with the decreasing distance $Dist_w$
between $w$ of the target image and that of reconstructed image.
Intuitively, the image of one certain identity is usually composed of several facial attributes (e.g., the uniqueness combination of face edges, eyes and nose shape, etc).
Thus, we further conduct the classification on five facial attributes and identity recognition tasks, both of which showing the same trending changes to the target identity. 
This is reasonable since facial attributes should change gradually within a continuous face manifold.
By slightly changing the prediction values, we can gradually alter the semantic attributes of the corresponding identity (such as the zoom-in parts of mouth and eye).

\subsection{Training Data Selection}
Since private data is unavailable during the whole procedure, we resort to the public dataset of same task as training data, which has no identity overlap with private dataset. 
Following \cite{yuan2023pseudo}, we input all public images into target model and obtain the prediction vectors. 
For each identity, we select top-$n$ images with the highest prediction scores for training.
Meanwhile, we also apply the same selection on the synthesized data\cite{yao2022style}, forming the same number of top-$n$ images for each identity.  
Finally, training data $D_{pub}$ can be expressed as $D_{pub} = {\{({x}_i, {p}_i)\}}_{i=1}^{N}$.
Note that the selected training data preserves facial attribute overlaps between public and private images, implicitly formulating a better prior for the training phase.

During the training phase, we first introduce the pixel-wise $\mathcal{L}_{mse}$ loss which is commonly used in image reconstruction:
\begin{equation}
 \mathcal{L}_{mse}=\left\|G{{(E({p}))}}-{x}\right\|_{2}.
\end{equation}

In addition, we adopt the perceptual loss \cite{DBLP:conf/cvpr/ZhangIESW18} to constrain the perceptual similarity between the reconstructed and real images:
\begin{equation}
\mathcal{L}_{lpips}=\left\|F({G{{(E({p}))}})}-F({{x}})\right\|_{2},
\end{equation}
where $F$ is the feature extractor \cite{yao2022style}.
The multi-layer identity loss and face parsing loss are also introduced for identity consistency: 
\begin{equation}
\mathcal{L}_{id}=\sum_{j=1}^{5}(1-\left\langle\mathbf{R}_{j}(G{{(E({p}))}}),\mathbf{R}_{j}({x})\right \rangle),
\end{equation}
\begin{equation}
\mathcal{L}_{parse}=\sum_{j=1}^{5}(1-\left\langle\mathbf{P}_{j}(G{{(E({p}))}}),\mathbf{P}_{j}({x})\right \rangle),
\end{equation}
where $\mathbf{R}$ is the pre-trained ArcFace network \cite{deng2019arcface}, $\mathbf{P}$\footnote{https: //github.com/zllrunning/face-parsing.PyTorch} the pre-trained face parsing model, $j$ the $j$-th feature of the pre-trained model, and $\left\langle\right \rangle$ the cosine similarity.

To make the intermediate feature $f_E$ better adapt to the ResNet backbone, and to prevent our PAE encoder from meaninglessly overfitting, we further propose the alignment regularization loss formalized as:
\begin{equation}
\mathcal{L}_{align\_reg}=\left\|f_E-{x}\right\|_{2}.
\end{equation}

In a nutshell, the overall loss function can be summed up as follows:
\begin{equation}
\mathcal{L}_{recon}=\mathcal{L}_{mse}+\lambda_{1}\mathcal{L}_{lpips}+\lambda_{2}\mathcal{L}_{id}+\lambda_{3}\mathcal{L}_{parse},
\end{equation}
\begin{equation}
\mathcal{L}_{total}=\mathcal{L}_{recon}+\lambda_{4}\mathcal{L}_{align\_reg}.
\end{equation}

\subsection{Aligned Ensemble Attack}
Note that during the attack phase, only the target label (one-hot prediction vector) is available for the adversary.
Unfortunately, we empirically find that simply inputting one-hot prediction will lead to a dramatically poor performance (shown in Tab. \ref{table:6}).
To tackle this, we argue that if an image is classified as the target identity, it must contain at least partial characteristic facial attributes of this identity. 
This motivates us to ensemble target's attributes contained in different public images.
We thus propose the aligned ensemble attack to integrate the latent codes $w$ encoded by predictions $p$ for better reconstruction.
Specifically, given a target identity c, we perform the aligned ensemble as:
\begin{equation}
\mathcal{W}_{ens}=\frac{\sum_{i=1}^{n}{\max{{p}_i}\cdot{w_i}}}{\sum_{i=1}^{n}{\max{{p}_i}}},
\end{equation}
where $\mathcal{W}_{ens}$ denotes the ensembled latent code.  
Moreover, in light of the prediction-image alignment provided by our method, we can explicitly enhance the prediction vectors of target identity by the following interpolation: 
\begin{equation}
{S_c({{p}_i})}={S_c({{p}_i})}+m,
\end{equation}
where $m$ is the enhancement hyper-parameter of the increase on prediction value, $S_c(p_i)$ the prediction score on dimension $c$ of $p_i$.
To ensure the sum of prediction vector as $1$, we further adjust each non-target dimension by: ${S_k({{p}_i})}={S_k({{p}_i})}-{m\cdot{S_k({p}_i)}}/{(1-S_c({{p}_i}))}$, where $k\neq{c}$.
This is reasonable and consistent with the editability of StyleGAN inversion, i.e., allowing the change of corresponding attributes by manipulating a directional vector in the latent space, thus leading to further improvement on inversion performance. 

\section{Experiments}
\subsection{Settings}

\textbf{Datasets.}
We conduct experiments on three face benchmarks, i.e., CelebA \cite{liu2015deep}, FaceScrub \cite{ng2014data} and PubFig83 \cite{pinto2011scaling}. 
CelebA consists of 202,599 face images belonging to 10,177 identities, of which we use 30,027 images of 1,000 identities as the private dataset. 
FaceScrub consists of 106,863 face images with 530 identities, and 200 of them are randomly selected to form the private dataset. 
For PubFig83, 50 of the 83 identities are randomly selected to form the private dataset. 
All images are center cropped and resized to 64$\times$64.
Concretely, we use FFHQ \cite{karras2019style} and synthetic data \cite{yao2022style} as public dataset for training.

\noindent\textbf{Target models.}
For fair comparison, we adopt target models with different architectures: VGG16 \cite{DBLP:journals/corr/SimonyanZ14a}, ResNet-152 \cite{DBLP:conf/cvpr/HeZRS16} and Face.evoLVe \cite{DBLP:conf/iccvw/ChengZWXKSF17}, which are widely used backbones in previous methods \cite{zhang2020secret,chen2021knowledge,kahla2022label,han2023reinforcement}. 

\noindent\textbf{Evaluation metrics.}
Similar to \cite{han2023reinforcement}, we use the attack accuracy (Attack Acc), K-nearest neighbor distance (KNN Dist), feature distance (Feat Dist) and Learned Perceptual Image Patch Similarity (LPIPS) as evaluation metrics, see detailed information in supplementary material.

\noindent\textbf{Implementation details.}
The StyleGAN generator is pre-trained and fixed, while the PAE encoder is initialized with pre-trained parameters \cite{yao2022style}. 
We train our model for 30 epochs using the Ranger optimizer with an initial learning rate of $10^{-4}$, batch size 4, $\beta_1$ = 0.95 and $\beta_2$ = 0.999. 
We apply the log operator to the input prediction vectors to avoid the dominance effect \cite{DBLP:conf/ccs/ChenZSYH17} in the probability distribution. 
We set $\lambda_1=0.2$, $\lambda_2=0.1$, $\lambda_3=0.1$ and $\lambda_4=1$.
We train target model for 50 epochs using the SGD optimizer with an initial learning rate of 0.001, batch size 64, momentum 0.9 and weight decay $10^{-4}$.

\subsection{Comparison with SOTAs}
We compare our method with several state-of-the-art baselines across different settings.
Specifically, we include the white-box methods GMI \cite{zhang2020secret}, KED-MI \cite{chen2021knowledge}, LOMMA \cite{nguyen2023re} and PLG-MI \cite{yuan2023pseudo}, considered as the upper bounds.
We also implement black-box methods LB-MI \cite{DBLP:conf/ccs/YangZCL19}, MIRROR \cite{an2022mirror} and RLB-MI \cite{han2023reinforcement}, as well as lable-only methods BREP-MI \cite{kahla2022label} and LOKT \cite{nguyen2024label}.

\noindent\textbf{Standard setting.}
We first evaluate our approach using the previous standard setting: the public and private data come from the same dataset, with no identity overlaps. 
Tab. \ref{table:1} shows the results of our method and other baselines. 
Clearly, our method consistently perform best under the same data distribution:
1) Our method achieves optimal attack performance on all three target models.
Our attack performance surpasses previous black-box method RLB-MI, with attack accuracy on three target models increasing by 8.5\%, 7.6\%, and 4.7\%, respectively. 
Besides, our method has significantly narrowed the gap with SOTA white-box methods.
2) Note that although original LOKT performs well, it requires a significant amount of query cost (shown in Tab. \ref{table:7}). 
To ensure a fair comparison, we limit its number of queries and re-implement it as LOKT*. 
Obviously, its attack performance has decreased dramatically. 
3) Our method is significantly ahead in KNN Dist and Feat Dist, which also indicates that our method reconstructs images that are visually closer to private datasets rather than fitting the evaluation model on Attack Acc.

\begin{table}[tb]
    \centering
    \caption{
    Attack performance on different target models trained on CelebA. 
    All public datasets come from CelebA and no identity overlap with private dataset.
    }
    \tabcolsep=0.30cm
    \resizebox{0.7\linewidth}{!}{
        \begin{tabular}{cccccc}
            \hline\noalign{\smallskip}
            Target model & Type & Method & Attack Acc$\uparrow$ & KNN Dist$\downarrow$ & Feat Dist$\downarrow$ \\
            \noalign{\smallskip}\hline\noalign{\smallskip}
            \multirow{11}{*}{\makecell{VGG16\\(88\%)}} & \multirow{4}{*}{White-box} & GMI & 0.185 & 1693.7 & 1615.7 \\
             & & KED-MI & 0.703 & 1363.7 & 1288.4 \\
             & & LOMMA & 0.903 & 1147.4 & ${-}$ \\
             & & PLG-MI & 0.970 & 1080.5 & 985.5 \\             
            \noalign{\smallskip}\cline{2-6}\noalign{\smallskip}
             & \multirow{4}{*}{Black-box} & LB-MI & 0.075 & 1778.7 & 1741.6 \\
             & & MIRROR & 0.413 & 1456.1 & 1367.5 \\
             & & RLB-MI & 0.659 & 1310.7 & 1214.7 \\
             & & \cellcolor{gray!20}\textbf{ours} & \cellcolor{gray!20}\textbf{0.715} & \cellcolor{gray!20}\textbf{1039.1} & \cellcolor{gray!20}\textbf{920.8} \\
            \noalign{\smallskip}\cline{2-6}\noalign{\smallskip}
             & \multirow{3}{*}{Label-only} & BREP-MI & 0.581 & 1347.4 & 1256.5 \\
             & & LOKT & 0.873 & 1246.7 & ${-}$ \\
             & & \ LOKT* & 0.450 & 1294.7 & 1230.8 \\
            \noalign{\smallskip}\hline\noalign{\smallskip}
            \multirow{11}{*}{\makecell{Resnet152\\(91\%)}} & \multirow{4}{*}{White-box} & GMI & 0.300 & 1594.1 & 1503.5 \\
             & & KED-MI & 0.765 & 1277.3 & 1184.6 \\
             & & LOMMA & 0.929 & 1138.6 & ${-}$ \\
             & & PLG-MI & 1.000 & 1016.5 & 910.2 \\
            \noalign{\smallskip}\cline{2-6}\noalign{\smallskip}
             & \multirow{4}{*}{Black-box} & LB-MI & 0.041 & 1800.9 & 1735.7 \\
             & & MIRROR & 0.570 & 1360.7 & 1263.8 \\
             & & RLB-MI & 0.804 & 1217.9 & 1108.2 \\
             & & \cellcolor{gray!20}\textbf{ours} & \cellcolor{gray!20}\textbf{0.865} & \cellcolor{gray!20}\textbf{1015.8} & \cellcolor{gray!20}\textbf{896.3} \\
            \noalign{\smallskip}\cline{2-6}\noalign{\smallskip}
             & \multirow{3}{*}{Label-only} & BREP-MI & 0.729 & 1277.5 & 1180.4 \\
             & & LOKT & 0.921 & 1206.8 & ${-}$ \\
             & & \ LOKT* & 0.490 & 1293.9 & 1236.0 \\
            \noalign{\smallskip}\hline\noalign{\smallskip}
            \multirow{11}{*}{\makecell{Face.evoLVe\\(89\%)}} & \multirow{4}{*}{White-box} & GMI & 0.254 & 1628.6 & 1541.7 \\
             & & KED-MI & 0.741 & 1350.8 & 1261.6 \\
             & & LOMMA & 0.932 & 1154.3 & ${-}$ \\
             & & PLG-MI & 0.990 & 1066.4 & 972.6 \\             
            \noalign{\smallskip}\cline{2-6}\noalign{\smallskip}
             & \multirow{4}{*}{Black-box} & LB-MI & 0.111 & 1776.4 & 1729.1 \\
             & & MIRROR & 0.530 & 1379.7 & 1280.1 \\
             & & RLB-MI & 0.793 & 1225.6 & 1112.1 \\
             & & \cellcolor{gray!20}\textbf{our} & \cellcolor{gray!20}\textbf{0.830} & \cellcolor{gray!20}\textbf{974.1} & \cellcolor{gray!20}\textbf{862.4} \\
            \noalign{\smallskip}\cline{2-6}\noalign{\smallskip}
             & \multirow{3}{*}{Label-only} & BREP-MI & 0.721 & 1267.3 & 1164.0 \\
             & & LOKT & 0.939 & 1181.7 & ${-}$ \\
             & & \ LOKT* & 0.600 & 1262.0 & 1185.4 \\
            \hline
        \end{tabular}}
     
    \label{table:1}
\end{table}

\noindent\textbf{Distribution shifts.}
We also consider a more practical setting where the public dataset and private dataset are from different distributions.
Generally, as reported in Tab. \ref{table:2}, we can find that:
1) Our method achieves optimal performance on the distribution transfer among three datasets.
For instance, when attacking the target model trained on PubFig83, the attack accuracy reaches $82\%$, which is $32\%$ higher than second best black-box method RLB-MI, and even exceeds the white-box method KED-MI. 
We believe that the disentangled $\mathcal{W}^+$ space of StyleGAN allows the prediction vector representing the target identity with more general features. 
2) The training-based method LB-MI shows both poor attack accuracy and visual quality, which also keeps in line with its shallow structure and less semantic facial features.
3) In addition, the fewer the number of identities in the private dataset, the higher the attack accuracy. 
We analyze that fewer identities lead to fewer facial features included, thus it is easy to select images with high confidence scores for all identities in public dataset.

\begin{table}[tb]
    \caption{
        Attack performance on target model Face.evoLVe\protect\footnotemark \ trained with PubFig83, FaceScrub and CelebA, respectively.
        }
        
    \centering
    \tabcolsep=0.30cm
    \resizebox{0.6\columnwidth}{!}{
        \begin{tabular}{ccccccc}
            \hline\noalign{\smallskip}
            Public$\rightarrow$Private & Type & Method & Attack Acc$\uparrow$ & KNN Dist$\downarrow$ & Feat Dist$\downarrow$ & LPIPS$\downarrow$ \\
            \noalign{\smallskip}\hline\noalign{\smallskip}
            \multirow{8}{*}{\makecell{FFHQ$\rightarrow$PubFig83\\ \qquad \quad (92\%)}} & \multirow{3}{*}{White-box} & GMI & 0.20 & 1536.4 & 1696.2 & 0.416 \\
             & & KED-MI & 0.63 & 1211.0 & 1353.1 & 0.368 \\
             & & PLG-MI & 0.91 & 1211.9 & 1296.9 & 0.370 \\
            \noalign{\smallskip}\cline{2-7}\noalign{\smallskip}
             & \multirow{4}{*}{Black-box} & LB-MI & 0.42 & 1392.1 & 1579.0 & 0.488 \\
             & & MIRROR & 0.48 & 1028.4 & 1195.4 & 0.325 \\
             & & RLB-MI & 0.62 & 1193.4 & 1340.3 & 0.340 \\
             & & \cellcolor{gray!20}\textbf{ours} & \cellcolor{gray!20}\textbf{0.82} & \cellcolor{gray!20}\textbf{840.7} & \cellcolor{gray!20}\textbf{992.3} & \cellcolor{gray!20}\textbf{0.268} \\
            \noalign{\smallskip}\cline{2-7}\noalign{\smallskip}
             & Label-only & BREP-MI & 0.42 & 1230.8 & 1399.5 & 0.347 \\
            \noalign{\smallskip}\hline\noalign{\smallskip}
            \multirow{8}{*}{\makecell{FFHQ$\rightarrow$FaceScrub\\ \qquad \quad (96\%)}} & \multirow{3}{*}{White-box} & GMI & 0.23 & 1585.1 & 1612.1 & 0.381 \\
             & & KED-MI & 0.43 & 1520.9 & 1546.5 & 0.381 \\
             & & PLG-MI & 0.70 & 1344.5 & 1353.6 & 0.394 \\
            \noalign{\smallskip}\cline{2-7}\noalign{\smallskip}
             & \multirow{4}{*}{Black-box} & LB-MI & 0.14 & 1530.9 & 1581.9 & 0.517 \\
             & & MIRROR & 0.40 & 1362.4 & 1367.9 & 0.266 \\
             & & RLB-MI & 0.49 & 1451.8 & 1452.6 & 0.339 \\
             & & \cellcolor{gray!20}\textbf{ours} & \cellcolor{gray!20}\textbf{0.59} & \cellcolor{gray!20}\textbf{1243.8} & \cellcolor{gray!20}\textbf{1256.0} & \cellcolor{gray!20}\textbf{0.246} \\
            \noalign{\smallskip}\cline{2-7}\noalign{\smallskip}
             & Label-only & BREP-MI & 0.28 & 1539.0 & 1565.6 & 0.332 \\
            \noalign{\smallskip}\hline\noalign{\smallskip}
            \multirow{8}{*}{\makecell{FFHQ$\rightarrow$CelebA\\ \qquad \quad (93\%)}} & \multirow{3}{*}{White-box} & GMI & 0.17 & 1648.4 & 1580.9 & 0.403 \\
             & & KED-MI & 0.34 & 1525.9 & 1460.8 & 0.379 \\
             & & PLG-MI & 0.85 & 1332.7 & 1262.5 & 0.351 \\
            \noalign{\smallskip}\cline{2-7}\noalign{\smallskip}
             & \multirow{4}{*}{Black-box} & LB-MI & 0.07 & 1660.9 & 1594.7 & 0.503 \\
             & & MIRROR & 0.40 & 1360.9 & 1267.6 & 0.282 \\
             & & RLB-MI & 0.33 & 1519.9 & 1443.4 & 0.356 \\
             & & \cellcolor{gray!20}\textbf{ours} & \cellcolor{gray!20}\textbf{0.49} & \cellcolor{gray!20}\textbf{1302.9} & \cellcolor{gray!20}\textbf{1210.1} & \cellcolor{gray!20}\textbf{0.271} \\
            \noalign{\smallskip}\cline{2-7}\noalign{\smallskip}
             & Label-only & BREP-MI & 0.33 & 1503.2 & 1428.9 & 0.350 \\
            \hline
        \end{tabular}}
        
        \label{table:2}
\end{table}
\begin{table}[!htb]
    \caption{
    Attack performance on different target models trained on CelebA with FFHQ as the public dataset.
    }
    \tabcolsep=0.30cm
    \centering
    \resizebox{0.6\linewidth}{!}{
        \begin{tabular}{ccccccc}
            \hline\noalign{\smallskip}
            Target model & Type & Method & Attack Acc$\uparrow$ & KNN Dist$\downarrow$ & Feat Dist$\downarrow$ & LPIPS$\downarrow$ \\
            \noalign{\smallskip}\hline\noalign{\smallskip}
            \multirow{8}{*}{\makecell{VGG16\\(88\%)}} & \multirow{2}{*}{White-box} & GMI & 0.07 & 1410.6 & 1326.5 & 0.392 \\
             & & KED-MI & 0.30 & 1363.7 & 1288.4 & 0.378 \\
             & & PLG-MI & 0.86 & 1256.4 & 1162.0 & 0.336 \\
            \noalign{\smallskip}\cline{2-7}\noalign{\smallskip}
             & \multirow{4}{*}{Black-box} & LB-MI & 0.01 & 1317.1 & 1379.0 & 0.563 \\
             & & MIRROR & 0.19 & 1281.2 & 1178.1 & 0.276 \\
             & & RLB-MI & 0.29 & 1368.2 & 1304.0 & 0.350 \\
             & & \cellcolor{gray!20}\textbf{ours} & \cellcolor{gray!20}\textbf{0.35} & \cellcolor{gray!20}\textbf{1238.1} & \cellcolor{gray!20}\textbf{1118.0} & \cellcolor{gray!20}\textbf{0.258} \\
            \noalign{\smallskip}\cline{2-7}\noalign{\smallskip}
             & Label-only & BREP-MI & 0.26 & 1367.0 & 1274.0 & 0.351 \\
            \noalign{\smallskip}\hline\noalign{\smallskip}
            \multirow{8}{*}{\makecell{Resnet152\\(91\%)}} & \multirow{2}{*}{White-box} & GMI & 0.15 & 1421.6 & 1341.9 & 0.395 \\
             & & KED-MI & 0.47 & 1326.3 & 1239.2 & 0.365 \\
             & & PLG-MI & 0.97 & 1158.4 & 1047.4 & 0.352 \\
            \noalign{\smallskip}\cline{2-7}\noalign{\smallskip}
             & \multirow{4}{*}{Black-box} & LB-MI & 0.01 & 1371.4 & 1324.7 & 0.468 \\
             & & MIRROR & 0.24 & 1265.3 & 1153.4 & 0.283 \\
             & & RLB-MI & 0.39 & 1352.6 & 1259.7 & 0.359 \\
             & & \cellcolor{gray!20}\textbf{ours} & \cellcolor{gray!20}\textbf{0.46} & \cellcolor{gray!20}\textbf{1190.9} & \cellcolor{gray!20}\textbf{1073.3} & \cellcolor{gray!20}\textbf{0.253} \\
            \noalign{\smallskip}\cline{2-7}\noalign{\smallskip}
             & Label-only & BREP-MI & 0.38 & 1351.1 & 1271.9 & 0.354 \\
            \noalign{\smallskip}\hline\noalign{\smallskip}
            \multirow{8}{*}{\makecell{Face.evoLVe\\(89\%)}} & \multirow{2}{*}{White-box} & GMI & 0.14 & 1460.1 & 1356.5 & 0.402 \\
             & & KED-MI & 0.44 & 1316.4 & 1231.1 & 0.370 \\
             & & PLG-MI & 0.94 & 1246.7 & 1154.2 & 0.368 \\
            \noalign{\smallskip}\cline{2-7}\noalign{\smallskip}
             & \multirow{4}{*}{Black-box} & LB-MI & 0.01 & 1645.4 & 1494.1 & 0.484 \\
             & & MIRROR & 0.19 & 1242.1 & 1150.2 & 0.285 \\
             & & RLB-MI & 0.41 & 1302.8 & 1218.4 & 0.337 \\
             & & \cellcolor{gray!20}\textbf{ours} & \cellcolor{gray!20}\textbf{0.50} & \cellcolor{gray!20}\textbf{1181.4} & \cellcolor{gray!20}\textbf{1080.7} & \cellcolor{gray!20}\textbf{0.254} \\
            \noalign{\smallskip}\cline{2-7}\noalign{\smallskip}
             & Label-only & BREP-MI & 0.41 & 1346.4 & 1243.2 & 0.356 \\
            \hline
        \end{tabular}}
     
    \label{table:3}
\end{table}

\noindent\textbf{Different models.}
Tab. \ref{table:3} shows the results of attacking different model architectures with the target and evaluation models provided in \cite{chen2021knowledge,yuan2023pseudo}. 
Based on the comparisons, we observe that:
1) Clearly, for all model architectures, the attack accuracy of our method is consistently higher than other baselines. 
2) Surprisingly, the attack accuracy is almost $40$ times higher than the training-based LB-MI method, and an improvement of $18$\% compared to RLB-MI. 
This is reasonable due to the alignment of input predictions and disentangled latent codes.
3) Moreover, our method almost rivals the white-box method KED-MI, and even surpasses all the other methods on all the visual metrics.

\footnotetext{Note that in the previous works, no checkpoints of Face.evoLVe trained with PubFig83 or FaceScrub are provided. Thus we use three target models trained by ourselves according to previous methods in Tab. \ref{table:2}.}
\subsection{Ablation Study} \label{sec:4.3}
\textbf{Effect of loss terms.}
To evaluate the contribution of proposed loss terms, we train the model by removing each component solely and present the comparison results in Tab. \ref{table:4}.

As can be seen:
1) Configuration (Cfg.) A removes pixel reconstruction loss $\mathcal{L}_{mse}$ during training, which leads to less effective attack.
This can be owing to the disentanglement of learned facial features.
2) The decrease of Cfg. B indicates that $\mathcal{L}_{align\_reg}$ promises feature $f_E$ a better adaptation to the ResNet backbone, as well as a better prediction alignment with $\mathcal{W}^+$ space.
3) Cfg. C verifies that LPIPS loss greatly improves the performance, especially the perceived quality of reconstructed images. 
While Cfg. D shows the performance drop when adding the multi-scale LPIPS loss, which is unsuitable for the inversion process from predictions to images.
4) Cfg. E and F demonstrate that identity loss and parsing loss can better preserve the original identity of reconstructed images, thus further improve their perceived quality.

\begin{table}[tb]
    \caption{
    Effect of different loss terms.
    }
    \tabcolsep=0.35cm
    \centering
    \resizebox{0.7\columnwidth}{!}{
        \begin{tabular}{cc|cccc}
            \hline\noalign{\smallskip}
             & Configuration & Attack Acc$\uparrow$ & KNN Dist$\downarrow$ & Feat Dist$\downarrow$ & LPIPS$\downarrow$ \\
            \noalign{\smallskip}\hline\noalign{\smallskip}
            A.& w/o $\mathcal{L}_{mse}$ & 0.42 & 1337.5 & 1246.6 & 0.264 \\
            B.& w/o $\mathcal{L}_{align\_reg}$ & 0.47 & 1299.4 & 1210.9 & 0.266 \\
            C.& w/o $\mathcal{L}_{lpips}$ & 0.44 & 1333.6 & 1250.4 & 0.278 \\
            D.& w/ $\mathcal{L}_{m\_lpips}$ & 0.39 & 1295.6 & 1212.8 & 0.271 \\
            E.& w/o $\mathcal{L}_{id}$ & 0.03 & 1580.0 & 1499.8 & 0.263 \\
            F.& w/o $\mathcal{L}_{parse}$ & 0.46 & 1324.2 & 1246.7 & 0.268 \\
            \noalign{\smallskip}\hline\noalign{\smallskip}
            & our baseline & 0.49 & 1302.9 & 1210.1 & 0.271 \\
            \hline
        \end{tabular}}
    
    \label{table:4}
\end{table}
\begin{table}[tb]
    \caption{
        Ablation study on the proposed PAE encoder.
        }
    \centering
    \tabcolsep=0.35cm
    \resizebox{0.7\columnwidth}{!}{
        \begin{tabular}{cc|cccc}
            \hline\noalign{\smallskip}
             & Configuration & Attack Acc$\uparrow$ & KNN Dist$\downarrow$ & Feat Dist$\downarrow$ & LPIPS$\downarrow$ \\
            \noalign{\smallskip}\hline\noalign{\smallskip}
            G.& w/ $\mathcal{F}$ branch & 0.45 & 1285.4 & 1202.6 & 0.264 \\
            H.& Random initialization & 0.44 & 1303.9 & 1221.9 & 0.266 \\
            I.& replace with FC & 0.45 & 1322.6 & 1246.2 & 0.266 \\
            \noalign{\smallskip}\hline\noalign{\smallskip}
            & our baseline & 0.49 & 1302.9 & 1210.1 & 0.271 \\
            \hline
        \end{tabular}}   
        
    \label{table:5}
\end{table}

\noindent\textbf{Analysis on prediction alignment encoder.}
We comprehensively investigate several mechanisms of the proposed encoder and show the results in Tab. \ref{table:5}.
Specifically, Cfg. G adds the feature prediction $\mathcal{F}$ branch to our baseline and results in a decrease in attack accuracy.
In \cite{yao2022style}, the $\mathcal{F}$ branch directly replaces a certain layer of the generator in StyleGAN with the learned feature tensor, while will inevitably disrupt our inversion procedure.
Besides, Cfg. H initializes the image encoder randomly, achieving slightly better perceptual quality but worse attack performance.
This is acceptable since the pre-trained encoder have some prior knowledge.
We also replace the full PAE encoder with fully connected layers (Cfg. I), and the performance decrease indicates the importance of prediction alignment's structure.

\noindent\textbf{Effect of aligned ensemble attack.}
To evaluate the effectiveness of our aligned ensemble attack, we first replace it with the LB-MI method (using one-hot prediction vector as input). 
As shown in Tab. \ref{table:6}, the performance of one-hot prediction as input is extremely poor. 
We believe this is because our PAE encoder maps the prediction vectors into a high-dimensional space.
Using one-hot vectors as input will lead to the significant loss of information. 
We also directly ensemble the prediction vectors instead of the latent codes $w$, showing an obvious performance drop from our method. 
We consider that this is due to the difference between the prediction vector space and $\mathcal{W}^+$ space. 
The ensemble of prediction vectors is merely aiming at finding an numerical average.
However, our aligned ensemble aims to enhance the feature information of target identity, leading to more complementary facial attributes and better reconstruction.

\begin{figure}[t]
  \centering
   \includegraphics[width=\linewidth]{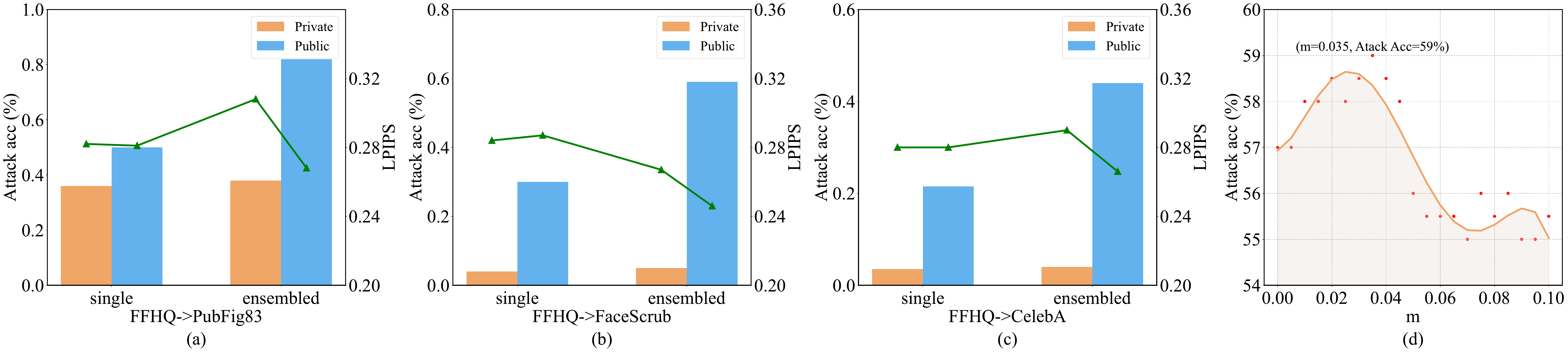}
   \caption{
   (a)-(c) show the result comparison of input predictions under single/ensembled and private/public settings.  
   (d) shows the sensitivity of hyper-parameter m.
   }
   \label{fig:4}
\end{figure}

In addition, Fig. \ref{fig:4}(a)-(c) show the attack performance of four variants on three private datasets, respectively, i.e., single private prediction, single public prediction, ensemble $w$ of private prediction and ensemble $w$ of public prediction.
Note that private prediction vector is only presented for thorough analysis, which is unavailable during the attack phase. 
Obviously, it is rather difficult to achieve high attack accuracy when using single prediction vector as input. 
We believe that reconstructing high-dimensional images from low-dimensional prediction vectors is inherently challenging. 
Luckily, the more disentangled nature of $\mathcal{W}^+$ space in StyleGAN makes our ensemble scheme an effective way to compensate for the aforementioned shortcomings. 
As shown in Fig. \ref{fig:4}(d), we compare the attack performance by increasing different values $m$ in the target dimension of public prediction vectors. 
When $m=0$, it means no modification is performed on the public prediction vectors. 
As $m$ increases, the target dimension value of public prediction vectors will also increase, along with the improved attack performance. 
However, if $m$ continues to increase (from 0.035 to 0.1), the value change on vectors may be so drastic that the altered prediction deviates from the original distribution, resulting in a significant performance decrease.

\begin{figure*}[t]
\parbox[t]{.48\linewidth}{
  \centering
  \captionof{table}{
    Comparison of attack performance by different input schemes.
    }
  \tabcolsep=0.35cm
  \resizebox{0.45\columnwidth}{!}{
        \begin{tabular}{cccc}
            \hline
            Attack Phase & PubFig83 & FaceScrub & CelebA \\
            \hline
            one-hot prediction & 0.02 & 0.005 & 0 \\
            prediction ensemble & 0.68 & 0.45 & 0.38 \\
            aligned ensemble (ours) & 0.82 & 0.59 & 0.49 \\
            \hline
        \end{tabular}
        }
    
    \label{table:6}
    }
    \hfill
    \parbox[t]{.5\linewidth}{
    \centering
    \captionof{table}{
    Comparison with other black-box methods on query number to target model. 
    }
    \tabcolsep=0.25cm
    \resizebox{0.5\columnwidth}{!}{
        \begin{tabular}{cccccc}
            \toprule
            Method&\textbf{ours}&RLB-MI&MIRROR&BREP-MI&LOKT\\
            \midrule
            Query number (million)&\textbf{0.13} & 36 & 1.85 & 17.98 & 12.66\\
            \midrule
            Percentage& ${-}$ & $0.36\%$ & $7.02\%$ & $0.72\%$ & $1.02\%$\\
            \bottomrule
           
        \end{tabular}
        }
     
    \label{table:7}
    }
\end{figure*}

\noindent\textbf{Query cost.}
Tab. \ref{table:7} shows the query costs of our method compared to others. We attack 300 identities in total. It is clear to see that the queries of ours is only 0.13 million, which is approximately only 0.36\% of RLB-MI.
In real-world scenarios, many MaaS platforms, such as Google and Amazon, limit the number of queries to the model, making black-box MI attack methods that rely on massive queries impractical. 
Our method only requires a small number of queries for data selection, taking an important step towards practical application of the black-box MI attack.

\begin{figure*}[t]
  \centering
   \includegraphics[width=\linewidth]{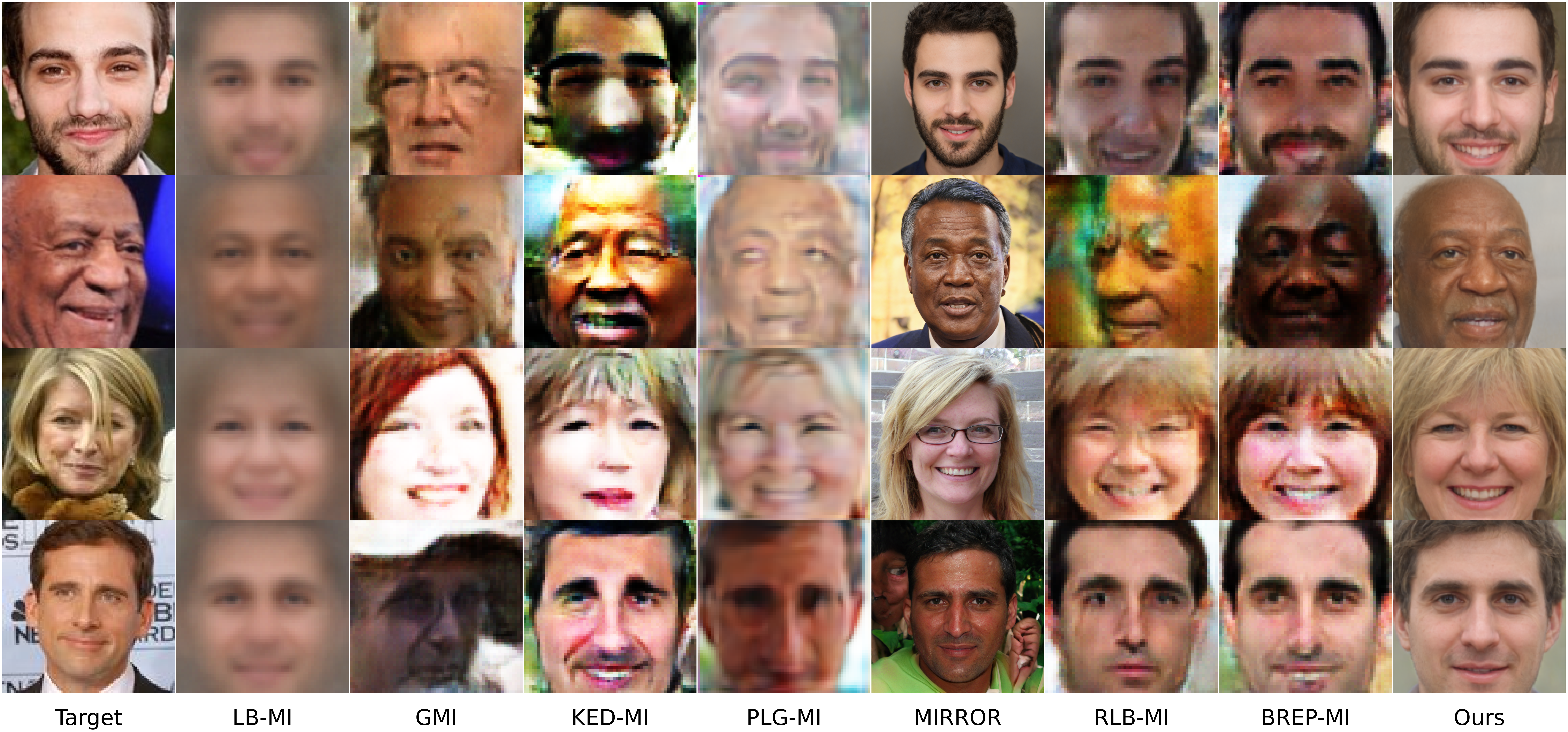}
    \caption{
    Visual comparison of different model inversion attacks.
    }
    \label{fig:5}
\end{figure*}

\noindent\textbf{Visualizations.}
Fig. \ref{fig:5} shows the qualitative results of different inversion methods. 
Compared with other baselines, our reconstructed images obviously are more realistic and have higher resolution quality, verifying that the alignment provides more characteristic facial features of the target identity.

\section{Conclusion}
We propose a novel training-based Prediction-to-Image (P2I) method for black-box model inversion attack. 
P2I maps the target model's output predictions into the more disentangled latent space of StyleGAN, providing alignment between predictions and reconstructed high-fidelity images.
We further design the aligned ensemble attack to enhance the feature information of target identity, leading to more complementary facial attributes and better reconstruction.
Extensive experiments show the effectiveness of our method on both attack performance and visual quality.
This work highlights that the rich information hidden in the model's prediction can be extracted, leading to data privacy leakage.
We hope this will raise the attention of the community on facial privacy protection.

\textbf{Limitations.}
One limitation is that we have not fully explored the potential of this StyleGAN-based training paradigm in model inversion task. 
In the future, we will continue to explore the essence of latent space in model inversion attack to further improve the performance of attacks in black-box or label-only scenario.

\textbf{Negative social impacts.}
Our work might be adopted by malicious users to expose the privacy leaks that exist in online models.
However, in light of this work, we hope to explore effective defense methods to counter different types of MI attacks, mitigating the underlying negative impact.


\section*{Acknowledgements}
This work was supported by the National Key R\&D Program of China under Grant 2022YFB3103500, the National Natural Science Foundation of China under Grants 62202459 and 62106258, and Young Elite Scientists Sponsorship Program by CAST (2023QNRC001) and BAST (NO.BYESS2023304).

%
%
\bibliographystyle{splncs04}
\bibliography{main}
\clearpage
\renewcommand\thesection{\Alph{section}}
\setcounter{section}{0}
\section*{Appendix}
\section{Evaluation metrics details}
\label{sec:a}
The detailed description of the evaluation metrics we use is as follows:

\textbf{Attack Acc} evaluates the accuracy of the reconstructed image on the target identity via an evaluation classifier, considered as the replacement of human judgment. 
We use the model proposed by Cheng \etal \cite{DBLP:conf/iccvw/ChengZWXKSF17} for evaluation, which is pre-trained on MS-Celeb-1M \cite{DBLP:conf/eccv/GuoZHHG16} and fine-tuned on the private dataset. 

\textbf{KNN Dist} is the average $L_2$ distance of reconstructed image features and features of the nearest image corresponding to target identity.

\textbf{Feat Dist} is the average $L_2$ distance of the reconstructed image features and the centroid of image features corresponding to the target identity. 
Both of them are calculated with features of the layer before fully connected layer in classifier.

\textbf{LPIPS} evaluates the perceptual similarity of reconstructed image and target identity image, which is more in line with human perception.

\section{More analyses on training data}
\label{sec:b}
In order to study the impact of training data on our method, we first assess the impact of synthesized data in the public dataset under the standard setting. 
The results in Tab. \ref{table:8} show that the lack of synthesized data lead to a clear decline in attack performance although slightly better perceptual quality. 
As the previous analysis indicates in \cite{yao2022style}, synthesized data is generated from real latent codes in the $\mathcal{W}^+$ space, which helps our PAE to align the prediction vector space with the $\mathcal{W}^+$ space. 
However, real latent codes might be outside the real image domain, thereby causing a slight impact on the LPIPS.

In addition, we randomly select the same amount of data from the FFHQ dataset and synthesized data, instead of our top-$n$ selection from public dataset, to train our encoder. 
The results in Tab. \ref{table:9} show that training our encoder with unselected public data results in a $42\%$ decrease from our baseline method.
We believe that unselected training data fails to allow our encoder to learn a good mapping relationship from prediction vectors to $\mathcal{W}^+$ space. 
In other words, since public images do not belong to any target identity, there may exist some images that have rather even prediction values on each identity, exerting negative effect on optimizing our PAE encoder.

\begin{table}[thb]
    \centering
    \tabcolsep=0.30cm
    \caption{
    Attack performance when synthesized data is not in the public dataset. Target model is VGG16 trained on CelebA. Public dataset also comes from CelebA, whereas there is no identity overlap with the private dataset.
    } 
    \resizebox{0.6\linewidth}{!}{
        \begin{tabular}{ccccc}
            \hline\noalign{\smallskip}
            Type& Attack Acc$\uparrow$ & KNN Dist$\downarrow$ & Feat Dist$\downarrow$ & LPIPS$\downarrow$ \\
            \noalign{\smallskip}\hline\noalign{\smallskip}
            w/o synthesized data & 0.705 & 1081.5 & 973.3 & 0.241 \\
            \cellcolor{gray!20}\textbf{ours} & \cellcolor{gray!20}\textbf{0.715} & \cellcolor{gray!20}\textbf{1039.1} & \cellcolor{gray!20}\textbf{920.8} & \cellcolor{gray!20}0.244 \\
            \noalign{\smallskip}\hline
        \end{tabular}}
    \label{table:8}
\end{table}

\section{More analyses on aligned ensemble attack}
\label{sec:c}
\begin{figure*}[ht]
  \centering
   \includegraphics[width=0.5\linewidth]{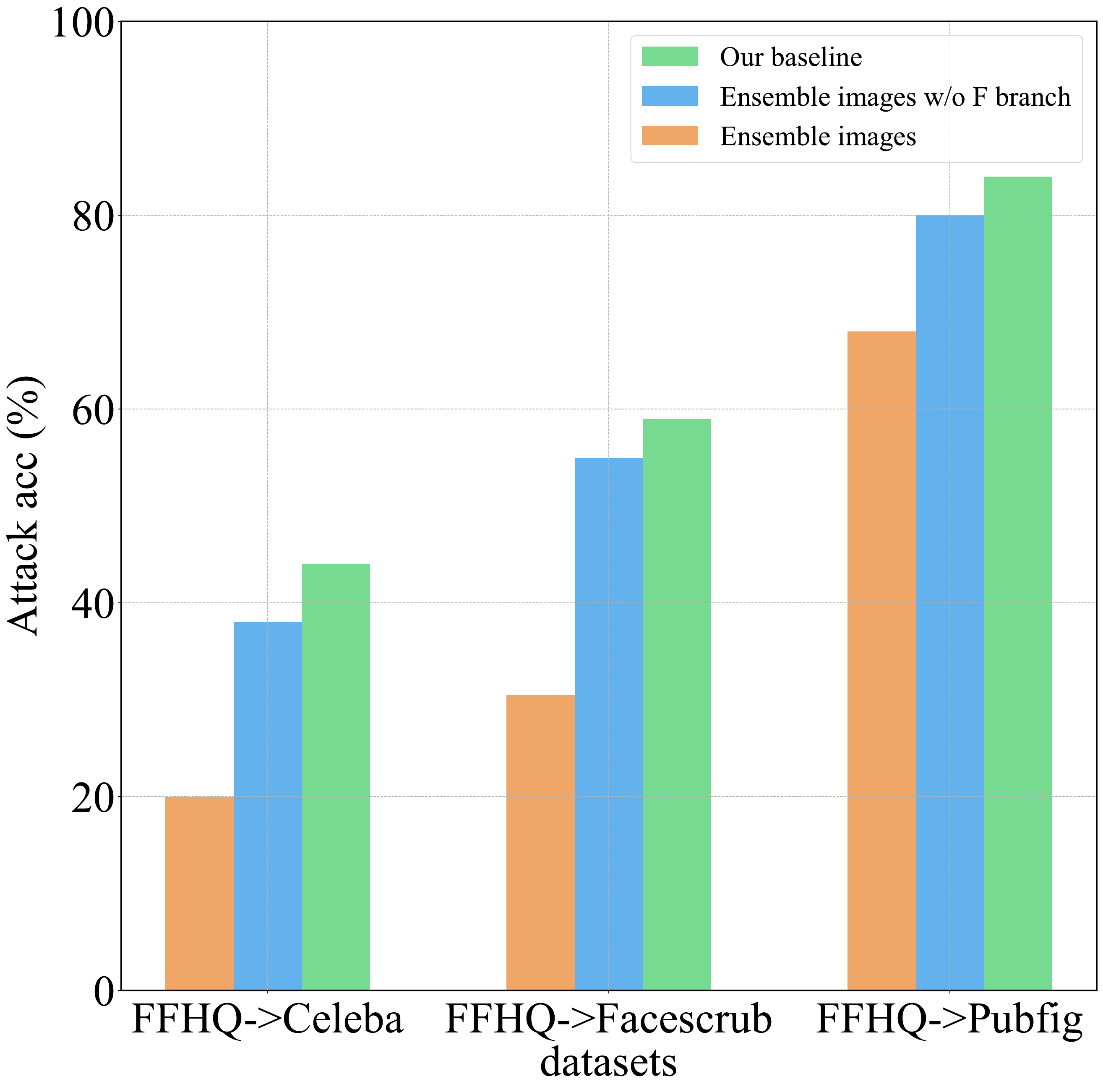}
    \caption{
    Comparison on aligned ensemble attack between ensembled latent codes of prediction vectors and those of images.
    }
    \label{fig:6}
\end{figure*}

We compare our baseline with the variant of adapting aligned ensemble attack on images' latent codes, as shown Fig. \ref{fig:6}.
Specifically, we directly feed the set of top-$n$ public images, classified as the target identity, into the pre-trained image encoder to reconstruct images \cite{yao2022style}. 
As can be seen, if we weight both the latent code in the $\mathcal{W}^+$ space and the feature tensor from $\mathcal{F}$, the attack performance becomes decreased as shown by the orange bar vs. green bar. 
One possible reason is that weighted ensembling of feature tensors would compromise the integrity of the features. 
We also keep the input unchanged and only ensemble the latent code in the $\mathcal{W}^+$ space. 
The results show an improvement in attack accuracy, but there is still a gap with the baseline, as shown by the blue bar vs. green bar. 

\begin{table}[thb]
    \centering
    \tabcolsep=0.30cm
    \caption{
    Attack performance when there is not data selection in our method. 
    The target model is Face.evoLVe trained on FaceScrub with FFHQ as the public dataset. The test accuracy of the target model is 96\%.
    } 
    \resizebox{0.6\linewidth}{!}{
        \begin{tabular}{ccccc}
            \hline\noalign{\smallskip}
            Type& Attack Acc$\uparrow$ & KNN Dist$\downarrow$ & Feat Dist$\downarrow$ & LPIPS$\downarrow$ \\
            \noalign{\smallskip}\hline\noalign{\smallskip}
            w/o data selection & 0.34 & 1294.3 & 1290.5 & 0.249 \\
            \cellcolor{gray!20}\textbf{ours} & \cellcolor{gray!20}\textbf{0.59} & \cellcolor{gray!20}\textbf{1243.8} & \cellcolor{gray!20}\textbf{1256.0} & \cellcolor{gray!20}\textbf{0.246} \\
            
            \noalign{\smallskip}\hline
        \end{tabular}}
    \label{table:9}
\end{table}

\section{Ablation on StyleGAN}
\label{sec:d}
To further study the impact of StyleGAN in our method, we compare two variants with our baseline. 
On the one hand, we replace the StyleGAN in our pipeline with the conventional GAN in RLB-MI \cite{han2023reinforcement}.
To keep consistency with our pipeline, we only add a linear layer after our PAE encoder to align dimensions. 
On the other hand, we switch our PAE encoder to one single linear layer, and replace $\mathcal{L}_{total}$ with $\mathcal{L}_{mse}$. 
Also, during the attack phase, we use one-hot vectors as input instead of our aligned ensemble attack. 
By doing this, we  only replace the generative network of LB-MI \cite{DBLP:conf/ccs/YangZCL19} with that of StyleGAN, do the minimal modifications for adjusting the pipeline, and keep other parts unchanged. 

From the results of Tab. \ref{table:10}, we can draw the following three conclusions: 

1. Due to the inferior priors of a regular GAN compared to StyleGAN, and its lack of the disentangled latent space, it performs worse than StyleGAN in the same pipeline. 

2. For the LB-MI pipeline, using the more powerful StyleGAN does not improve the attack performance. 
We believe that in the LB-MI pipeline, there is a lack of a good encoder to map prediction vectors into the latent space of the generator, thus the powerful prior of StyleGAN cannot be fully utilized. 
Also, the one-hot vector leads to too much loss of information, which is insufficient for high-fidelity reconstructions. 

3. Comparing two variants with the results from our baseline, it underlines that the key characteristics of our pipeline and StyleGAN are complementing each other. Concretely, StyleGAN possesses strong prior knowledge. 
Meanwhile, our specially designed PAE encoder can effectively utilize StyleGAN's disentangled space.
Moreover, our loss function is better equipped to guide the alignment of the prediction vector space and $\mathcal{W}^+$ space. 
During the attack phase, our proposed alignment integrated attack is also based on the disentangled characteristics of StyleGAN's $\mathcal{W}^+$ space.



\begin{table}[h]
    \centering
    \tabcolsep=0.30cm
    \caption{
   Ablation on StyleGAN. 
   The target model is VGG16 trained on CelebA (also as the public dataset). 
   \textbf{Variant1}: replace the StyleGAN in our pipeline with the conventional GAN in RLB-MI. 
   \textbf{Variant2}: combine the StyleGAN with LB-MI.
    } 
    \resizebox{0.6\linewidth}{!}{
        \begin{tabular}{ccccc}
            \hline\noalign{\smallskip}
            Type& Attack Acc$\uparrow$ & KNN Dist$\downarrow$ & Feat Dist$\downarrow$ & LPIPS$\downarrow$ \\
            \noalign{\smallskip}\hline\noalign{\smallskip}
            variant1 & 0.125 & 1629.0 & 1565.1 & 0.304 \\
            variant2 & 0.005 & 2347.7 & 2306.2 & 0.527 \\
            \cellcolor{gray!20}\textbf{ours} & \cellcolor{gray!20}\textbf{0.715} & \cellcolor{gray!20}\textbf{1039.1} & \cellcolor{gray!20}\textbf{920.8} & \cellcolor{gray!20}\textbf{0.244} \\
            
            \noalign{\smallskip}\hline
        \end{tabular}}
    
    \label{table:10}
\end{table}

\section{User study}
\label{sec:e}
\begin{figure*}[h]
  \centering
   \includegraphics[width=\linewidth]{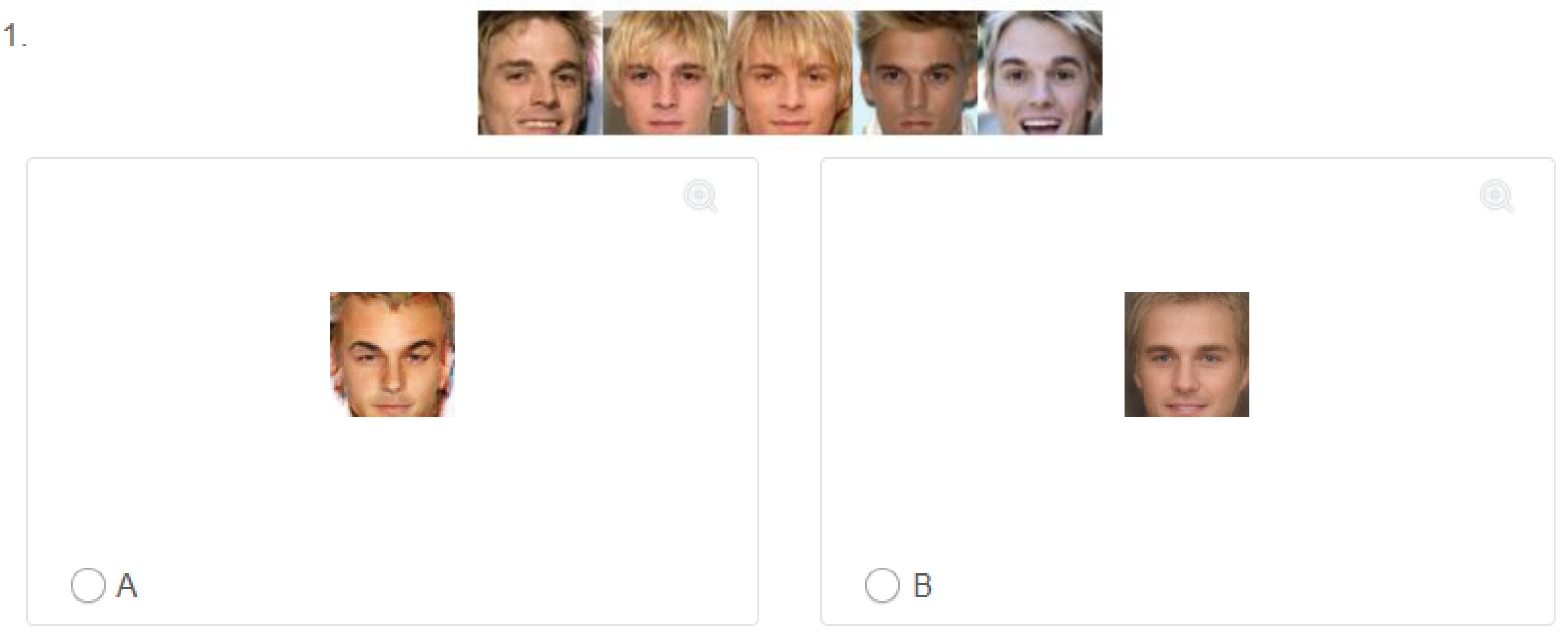}
    \caption{
    User study interface
    }
    \label{fig:7}
\end{figure*}

To further quantify the accuracy of the images reconstructed using our method, we introduce human study to evaluate the effectiveness of the attack from human subjective perspective. 
We design our experiment according to the setup in previous work \cite{nguyen2024label}. 
As shown in Fig. \ref{fig:7}, we randomly select five real images for each target identity from the private dataset to serve as references for the users. 
The users need to choose between two options, deciding which one image matches the reference identity better. 
The two options are the images obtained from attacks on the reference identity using RLB-MI and our method respectively. The target model is Face.evoLVe
trained on CelebA (also as the public dataset). 
We randomly select 50 identities and hand them over to 20 users for experimentation.

The results in Tab. \ref{table:11} show that user preferences for the inversion results of our method reach 74.40\%, while RLB-MI only with 25.60\%. 
This result indicates that users subjectively prefer our results, therefore our method also has superiority in visual quality.

\begin{table}[!t]
    \centering
    \tabcolsep=0.30cm
    \caption{
      User study results. User preferences indicate which option users prefer between two methods of inversion images. The results show that users prefer the outcomes produced by our method.
    } 
    \resizebox{0.35\linewidth}{!}{
        \begin{tabular}{cc}
            \toprule
            Method& User Preference$\uparrow$ \\
            \midrule
            RLB-MI & 25.60\% \\
            \cellcolor{gray!20}\textbf{ours} & \cellcolor{gray!20}\textbf{74.40\%} \\
            \bottomrule
        \end{tabular}}
    
    \label{table:11}
\end{table}

\begin{table}[thb]
    \renewcommand{\arraystretch}{1.2}
    \centering
    \caption{
        Comparison of implicit attribute recovery. 
        We compare the state-of-the-art methods with our method, and the attack success rate is measured by the attribute classifier trained on CelebA.
        }
    \resizebox{0.5\columnwidth}{!}{
        \begin{tabular}{ccccc}
            \toprule
            \multirow{2}{*}{Attributes} & \multicolumn{4}{c}{Attack Acc$\uparrow$}\\
            \cmidrule(l){2-5}
             & PLG-MI & RLB-MI & BREP-MI & \textbf{ours} \\
            \midrule
            Bald & 0.985 & 0.982 & 0.980 & \textbf{0.986} \\
            Big Lips & 0.424 & 0.421 & 0.373 & \textbf{0.435} \\
            Brown Hair & 0.516 & 0.509 & 0.451 & \textbf{0.565} \\
            Chubby & 0.876 & 0.825 & 0.882 & \textbf{0.884} \\
            Double Chin & 0.873 & 0.877 & 0.863 & \textbf{0.884} \\
            Eyeglasses & 0.905 & 0.877 & 0.922 & \textbf{0.928} \\
            Gray Hair & 0.935 & 0.895 & 0.941 & \textbf{0.942} \\
            Heavy Makeup & 0.517 & 0.439 & 0.451 & \textbf{0.551} \\
            High Cheekbones & 0.667 & 0.667 & 0.725 & \textbf{0.899} \\
            Narrow Eyes & 0.902 & 0.930 & 0.882 & \textbf{0.942} \\
            Wearing Necktie & 0.772 & 0.825 & 0.824 & \textbf{0.855} \\
            \midrule
            All & 0.695 & 0.702 & 0.712 & \textbf{0.726} \\
            \bottomrule
        \end{tabular}}
        
        \label{table:12}
\end{table}

\section{Attributes recovery}
\label{sec:f}
In this section, we evaluate the performance of our method in attribute restoration to further verify the effectiveness of image reconstruction. 
Specifically, we train 40 attribute classifiers on the CelebA dataset. 
Since different images of the same identity may have different attribute features, it is reasonable that the same identity may be different due to different occasions, dress ups and other factors. 
We take the union of the attributes of different images belonging to the same identity, to form the feature attribute set for calculating the performance of attribute restoration. 
We calculate the attack accuracy of identity feature on images that are successfully attacked using different methods.

The results in Tab. \ref{table:12} show that our method is the best in terms of attack performance on all 40 attributes of the CelebA dataset, with an improvement of $3.4\%$ over the black-box method RLB-MI and even $4.5\%$ higher than the white-box method PLG-MI. 
We also select some of these attributes and calculate the attack accuracy on individual attributes. 
On the selected individual attributes, our attack performance is still higher than that of other methods.
These experimental results demonstrate the advantages of our method in attribute restoration, further proving that our PAE encoder aligns the prediction vector space with the more disentangled $\mathcal{W}^+$ space. 
Furthermore, the aligned ensemble attack contributes to aggregate different feature attributes of the same identity together.

\section{Effect of attribute editing on MI attack}
\label{sec:g}
\begin{figure}[h]
\centering
  \includegraphics[width=0.7\linewidth]{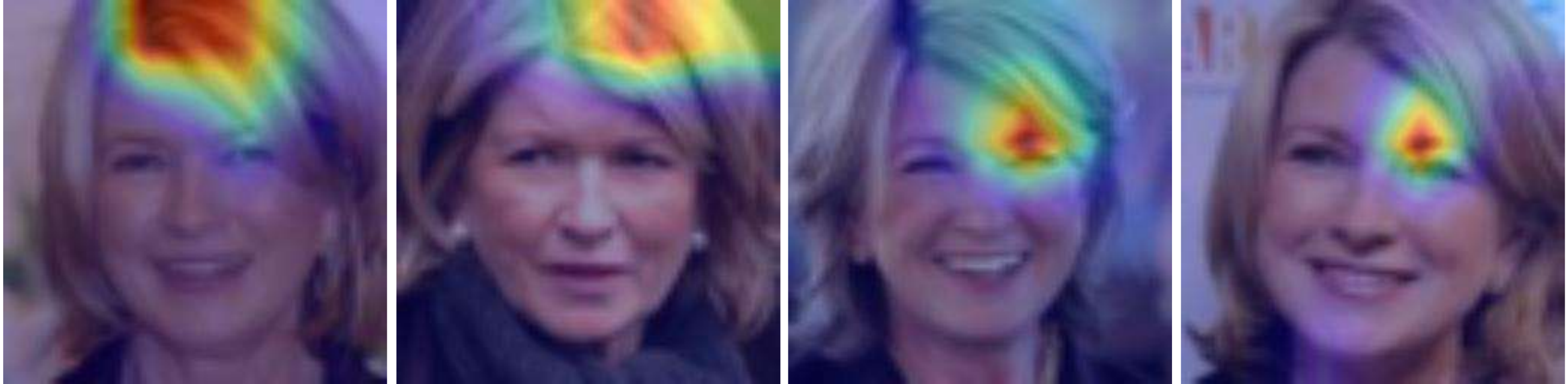}
   \caption{Heatmap of target identity on target model.}
   \label{fig:8}
\end{figure}

We also explore the potential for more targeted manipulation of facial attributes within $\mathcal{W}^+$ space. To integrate attribute editing into the inversion process, given a target identity, we obtain its latent code $\mathcal{W}_{ens}$, visualize its heatmap using Grad-CAM, and utilize InterFaceGAN \cite{shen2020interpreting} to find editing directions in $\mathcal{W}^+$ space of StyleGAN2.
As in Figure \ref{fig:8}, we select two editing directions according to the heatmap: $\mathcal{W}_{golden\_hair}$ and $\mathcal{W}_{bushy\_eyebrows}$. 
Thus, we can conduct the interpolation as: $\mathcal{W}_{edit}$=$\mathcal{W}_{ens}$+$\alpha$$\mathcal{W}_{golden\_hair}$(or $\mathcal{W}_{bushy\_eyebrows}$) to edit $\mathcal{W}_{ens}$.

\begin{figure}[h]
\centering
  \includegraphics[width=\linewidth]{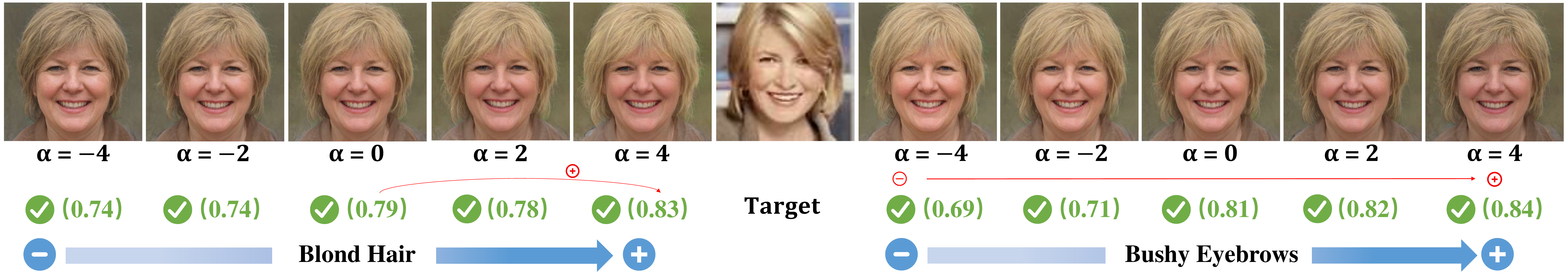}
   \caption{We attack the target identity and manipulate the obtained $\mathcal{W}_{ens}$ to edit attributes.}
   \label{fig:9}
\end{figure}

Figure \ref{fig:9} shows the effect of attribute editing on Model Inversion attack: 
1) Even though our PAE encoder is trained using the prediction vectors as input, we can still effectively conduct edits using the editing directions obtained from the image-to-image reconstruction pipeline \cite{shen2020interpreting}. 
This \textit{coincidentally} proves that our encoder has achieved alignment between the predictions and the $\mathcal{W}^+$ space. 
2) We are astonished to discover that manipulating latent code in a specific direction indeed boosts the classification accuracy of inversion images in our pipeline. 
This idea holds immense potential for further improving the efficacy of our method.
3) In the MI attacks, we argue that manipulating facial attributes still needs to address two problems: What are the properties closely related to the classification? How to determine the amount of attribute editing to improve the attack accuracy? We will explore these in the future.

\end{document}